%% file: acl_latex.tex
\pdfoutput=1

\documentclass[11pt]{article}

\usepackage[preprint]{acl}

\usepackage{times}
\usepackage{latexsym}

\usepackage[T1]{fontenc}

\usepackage[utf8]{inputenc}

\usepackage{microtype}

\usepackage{inconsolata}

\usepackage{graphicx}
\usepackage{amssymb}
\usepackage{algorithm}
\usepackage{algpseudocode}
\usepackage{amsmath}
\usepackage{textcomp}
\usepackage{booktabs}
\usepackage{multirow}
\title{Speculative Decoding for Multi-Sample Inference}

\author {
    \textbf{Yiwei Li}\textsuperscript{\rm 1}, \hspace{0cm}
    \textbf{Jiayi Shi}\textsuperscript{\rm 1}, \hspace{0cm}
    \textbf{Shaoxiong Feng}\textsuperscript{\rm 2}, \hspace{0cm} 
    \textbf{Peiwen Yuan}\textsuperscript{\rm 1}, \hspace{0cm} 
    \textbf{Xinglin Wang}\textsuperscript{\rm 1}, \hspace{0cm} \\
    \textbf{Yueqi Zhang}\textsuperscript{\rm 1}, \hspace{0cm} 
    \textbf{Ji Zhang}\textsuperscript{\rm 1}, \hspace{0cm} 
    \textbf{Chuyi Tan}\textsuperscript{\rm 1}, \hspace{0cm} 
    \textbf{Boyuan Pan}\textsuperscript{\rm 2}, \hspace{0cm} 
    \textbf{Yao Hu}\textsuperscript{\rm 2}\textbf{,} \hspace{0cm} 
    \textbf{Kan Li}\textsuperscript{\rm 1}\footnotemark[2] \\
    \textsuperscript{\rm 1} School of Computer Science, Beijing Institute of Technology \\
    \textsuperscript{\rm 2} Xiaohongshu Inc \\
    \texttt{\{liyiwei,shijiayi,peiwenyuan,wangxinglin\}@bit.edu.cn} \\
    \texttt{\{shaoxiongfeng2023\}@gmail.com} \  \texttt{\{panboyuan,xiahou\}@xiaohongshu.com} \\
     \texttt{\{zhangyq,jizhang,tanchuyi,likan\}@bit.edu.cn} 
}

\begin{document}
\maketitle

\input{abstract}
\input{introduction}
\input{background}
\input{method}

\input{experiment}
\input{related_work}
\input{conclusion}
\input{limitations}

\bibliography{anthology,custom}

\input{appendix}

\end{document}

%% file: abstract.tex
\begin{abstract}
We propose a novel speculative decoding method tailored for multi-sample reasoning scenarios, such as self-consistency and Best-of-N sampling. Our method exploits the intrinsic consensus of parallel generation paths to synthesize high-quality draft tokens without requiring auxiliary models or external databases. By dynamically analyzing structural patterns across parallel reasoning paths through a probabilistic aggregation mechanism, it identifies consensus token sequences that align with the decoding distribution. Evaluations on mathematical reasoning benchmarks demonstrate a substantial improvement in draft acceptance rates over baselines, while reducing the latency in draft token construction. 
This work establishes a paradigm shift for efficient multi-sample inference, enabling seamless integration of speculative decoding with sampling-based reasoning techniques.
\end{abstract}

%% file: introduction.tex
\section{Introduction}
Nowadays, large language models (LLMs) increasingly rely on multi-sample aggregation strategies, such as majority voting (self-consistency) \citep{COT} and Best-of-N sampling \citep{bestn}, to enhance prediction accuracy in complex reasoning tasks. By generating and aggregating multiple candidate outputs, these methods mitigate individual sampling errors and improve task performance, particularly in mathematical reasoning \citep{lu-etal-2023-survey}. However, this benefit comes at a significant computational cost: generating multiple samples inherently prolongs inference latency, posing a critical challenge for real-world applications where efficiency is paramount.

To address latency issues, speculative decoding \citep{spe,spe2,spe3} has emerged as a promising acceleration method. By predicting draft tokens in advance and verifying them against the original LLM, this approach reduces the number of full autoregressive steps. However, existing methods depend on external draft models \citep{li-etal-2024-eagle} or retrieval-augmented databases \citep{he-etal-2024-rest} to generate drafts, introducing three critical limitations: First, draft tokens sourced externally struggle to fully align with the original LLM’s output distribution, which lowers token acceptance rates and diminishes the potential speedup. Second, querying these external modules or databases inevitably incurs additional latency. Third, relying on auxiliary resources incurs extra costs—training a draft model or maintaining a database require additional computational and storage overhead.

A critical observation motivates our work: multi-sample inference yields a rich reservoir of high-quality draft tokens. When generating multiple parallel reasoning paths (e.g., for majority voting), the LLM produces outputs that, while varied in expression, are all drawn from the same underlying distribution. Frequently, these outputs share common substeps—such as intermediate equations in math problems—while differing in ordering or phrasing. Importantly, because all paths stem from the same model, their substeps naturally mirror its output tendencies, unlike externally sourced drafts. This inherent alignment makes them ideal candidates for speculative decoding, provided we can effectively extract and aggregate consensus patterns.

In this paper, we introduce a novel speculative decoding method explicitly designed for multi-sample reasoning scenarios. Our approach begins by leveraging the overlapping token subsequences generated across parallel reasoning paths to build a dynamic draft token pool. At each inference step, the most recent tokens from any path are used as queries to retrieve matching prefixes from other paths via suffix-based matching. The tokens immediately following these matches are aggregated as candidate drafts. These candidates are then organized into a weighted directed acyclic graph (DAG), where the edge weights reflect transition probabilities derived from the LLM’s distribution. A confidence-weighted search is performed on this DAG to extract the highest-likelihood token sequence—prioritizing paths with strong agreement. This consensus-driven process yields draft tokens that closely align with the model's output distribution, ultimately accelerating inference by reducing the number of full autoregressive steps without the need for external modules or datastores.

We evaluate our method on two mathematical reasoning benchmarks (GSM8K \citep{GSM8K} and MATH \citep{MATH}) using two widely adopted LLMs (Llama3-8B-Instruct and Qwen2.5-7B-Instruct) under multi-sample inference settings. Our approach achieves much higher token acceptance rate at identical draft lengths compared to baselines, demonstrating superior alignment with the original model’s distribution. Crucially, the draft construction process incurs lower latency than baselines, as it eliminates costly database queries or auxiliary model inferences. 

%% file: background.tex
\section{Background}
\paragraph{Multi-Sample Inference}
Aggregation strategies aim to enhance reasoning reliability by generating and aggregating multiple candidate outputs. Let $Y=\{y_1,y_2,\ldots,y_N\}$ denote $N$ parallel samples drawn from a language model $p_{\theta}$. Two dominant paradigms include:
\begin{itemize}
    \item Self-Consistency: Selects the final answer via majority voting over $Y$:
    \begin{equation}
    \hat{y}=\arg\max_{a\in\mathcal{A}}\sum_{i=1}^N\mathbb{I}(a=y_i),
\end{equation}
where $\mathcal{A}$ is the answer space and $\mathbb{I(\cdot)}$ is the indicator function.
    \item Best-of-N: Scores each sample $y_i$ with a reward function $G(y_i|x)$ and selects the highest-ranked candidate:
    \begin{equation}
    \hat{y}=\arg\max_{y_i\in Y}G(y_i|x),
\end{equation}
where $G$ may quantify sequence likelihood, alignment with domain-specific heuristics, or external verification signals.
\end{itemize}
While aggregation techniques enhance accuracy, generating multiple samples increases latency, thereby motivating the development of acceleration methods.

\paragraph{Speculative Decoding} Speculative decoding reduces inference latency by predicting $K$-step draft tokens $\{\tilde{y}^1,\ldots,\tilde{y}^K\}$ in advance and verifying them via the original model $p_{\theta}$. Traditional approaches rely on an auxiliary draft model or a retrieval datastore $\mathrm{D}$ to propose drafts. For each step $t$:
\begin{itemize}
    \item Draft Proposal: Generate $\tilde{y}^t\sim q(\cdot|y^{1:t-1})$ or retrieve  $\tilde{y}^t$ from $\mathrm{D}$.
    \item Parallel Verification: Compute $p_\theta(\cdot|y^{1:t-1})$ for all $\tilde{y}^{1:t}$.
    \item Acceptance Check: Accept the longest prefix $\tau\leq K$ where $p_\theta(\tilde{y}^t|y^{1:t-1})\geqq(\tilde{y}^t|y^{1:t-1})$.
\end{itemize}

%% file: method.tex
\section{Method}
Our method transforms the inherent consensus of multi-sample reasoning paths into high-quality draft tokens through three key components: (1) dynamic construction of a draft pool via cross-path suffix searching, (2) fuzzy suffix matching to handle lexical variations and (3) consensus-driven sequence extraction. The entire process operates during parallel decoding, requiring no external models or datastores. Please refer to Appendix~\ref{app:agm} for algorithm.

\paragraph{Dynamic Draft Pool Construction}
During parallel generation of $N$ reasoning paths $\{y_1,y_2,\ldots,y_N\}$, we iteratively build a draft pool by identifying overlapping token subsequences across paths. For any partial sequence $y_i^{1:t}$ on the path $i$, we use its $k$-token suffix $y_i^{t-k+1:t}$ as a query to search for matching prefixes in other paths $\{y_j\}_{j\neq i}$. Matched prefixes' subsequent tokens are aggregated as draft candidates. Formally, the candidate set $\mathcal{C}_t$ at step $t$ is:
\begin{equation}
    \mathcal{C}_t=\bigcup_{j\neq i}\left\{y_j^{t^{\prime}:}\mid\exists t^{\prime}\leq t,y_j^{1:t^{\prime}}\text{ ends with }y_i^{t-k+1:t}\right\}.
\end{equation}
Candidates inherently align with $p_\theta$, as all paths are derived from the same model.

\paragraph{Fuzzy Suffix Matching}
To handle minor lexical variations (e.g., "x²" vs. "x\textasciicircum2" in mathematical steps), we extend exact suffix matching with edit distance tolerance $\epsilon$. For query suffix $s$, we retrieve all $s^{\prime}$ where:
\begin{equation}
    \delta(s,s^{\prime})\leq\epsilon,
\end{equation}
ensuring semantically equivalent tokens contribute to the draft pool.

\paragraph{Consensus-Driven Draft Extraction with DAG}
We organize candidates into a \textbf{directed acyclic graph (DAG)} $\mathcal{G}=(V,E)$, where nodes ($V$) represent unique tokens, and identical tokens that appear within the same layer (i.e., at the same time step) across different candidates are merged to form the graph.
Edges ($E$) encode transitions weighted by:
\begin{equation}
    w(u,v)=\alpha \cdot \sum_{i}p_\theta(v_i|u_i)+(1-\alpha)\cdot\mathrm{Freq}(u\to v),
\end{equation}
where $p_\theta(v_i|u_i)$ represents the model's probability of generating $v$ from $u$ in the $i$-th instance. We weight through generation probabilities to prevent the selection of low-probability tokens, thereby improving the accept rate.
We extract the optimal draft sequence by greedily selecting the token with the highest weight ($\sum_{u : (u, v) \in E} w(u,v)$) at each layer. The search terminates upon reaching either the maximum draft token length $L$ or a leaf node.

There are three advantages of DAG over tree structure:
\textbf{Compact Structure}: Merges shared tokens (e.g., common math operators) into single nodes.
\textbf{Multi-Path Support}: Captures divergent reasoning branches (e.g., "x+1" vs. "x-1") via multiple edges.
\textbf{Probabilistic Ranking}: Edge weights quantify consensus between paths.

\paragraph{Draft acceptance}
We follow \citet{he-etal-2024-rest}, where draft tokens are validated by comparing them with tokens sampled from the model's conditional distribution at each position. Accepted draft tokens are those that match the sampled tokens, and any mismatch leads to rejection of subsequent drafts. This method ensures that the generated sequences align with standard autoregressive generation without any loss of accuracy.

%% file: experiment.tex
\section{Experiment}
\subsection{Experimental Setup}
\paragraph{Datasets and Models}

We evaluate our method on two widely-used mathematical reasoning benchmarks: GSM8K \citep{GSM8K} and MATH \citep{MATH}. We test on Qwen2.5-7B-Instruct \citep{qwen} and Llama-3-8B-Instruct \citep{llama} models.
All experiments are conducted on a single NVIDIA A100 GPU. For simplicity, edit distance tolerance $\epsilon$ and $\alpha$ is set to 1.

\paragraph{Baselines} We compare our method with REST \citep{he-etal-2024-rest} and EAGLE-2 \citep{li-etal-2024-eagle}. REST retrieves draft tokens from a pre-built datastore. In this work, we built the datastore with NuminalMath-CoT dataset \citep{numina}. EAGLE-2 leverages both the token sequence and the hidden state sequence to sequentially predict subsequent draft tokens. 
When comparing performance, we adopt accept length as the evaluation metric, since baseline methods does not optimize inference time for batch size greater than one, making it incompatible with multi-sample inference. Please refer to Appendix~\ref{app:metric} for details. 

\subsection{Main Results}
\begin{table}[htbp]
    \centering
    \small 
    \begin{tabular}{l|ccccc}
    \toprule
    Draft Length  & 6 & 4 & 3 & 2 & 1 \\
    \midrule
    Model& \multicolumn{5}{c}{GSM8K} \\
       
    \midrule
    REST    & 0.56 & 0.49 & 0.41 & 0.25 & 0.06   \\
    EAGLE-2 & 0.97 & 0.87 & 0.70 & 0.65 & 0.52   \\
    Ours    & \textbf{1.53} & \textbf{1.46} & \textbf{1.35} & \textbf{1.12} & \textbf{0.71}  \\
    \midrule
     & \multicolumn{5}{c}{MATH} \\
    \midrule
    REST     & 0.77 & 0.59 & 0.44 & 0.23 & 0.06  \\
    EAGLE-2  & 0.98 & 0.87 & 0.69 & 0.68 & 0.52  \\
    Ours     & \textbf{1.89} & \textbf{1.63} & \textbf{1.47} & \textbf{1.17} & \textbf{0.72}  \\
    
    \bottomrule
    \end{tabular}
    \begin{tabular}{l|ccccc}
    \toprule
    Draft Length  & 6 & 4 & 3 & 2 & 1 \\
    \midrule
    Model& \multicolumn{5}{c}{GSM8K} \\
       
    \midrule
    REST    & 0.52 & 0.46 & 0.41 & 0.26 & 0.07   \\
    EAGLE-2 & 1.71 & 1.41 & 1.32 & \textbf{1.12} & \textbf{0.70}   \\
    Ours    & \textbf{2.17} & \textbf{1.79} & \textbf{1.53} & \textbf{1.12} & 0.66  \\
    \midrule
     & \multicolumn{5}{c}{MATH} \\
    \midrule
    REST     & 0.57 & 0.50 & 0.40 & 0.24 & 0.08  \\
    EAGLE-2  & 1.48 & 1.18 & 1.09 & 0.94 & 0.63  \\
    Ours     & \textbf{2.95} & \textbf{2.15} & \textbf{1.76} & \textbf{1.26} & \textbf{0.71}  \\
    
    \bottomrule
    \end{tabular}

    \caption{Accpet length of Qwen2.5-7B-Instruct (up) and Llama3-8B-Instruct (down) under different draft length $L$. Our method can generally accept more tokens.}
    \label{tab:result-main}
\end{table}

\paragraph{Draft Tokens Acceptation}
Table~\ref{tab:result-main} presents the accept length of our method and the baseline approaches under different draft lengths, where our method demonstrates a clear advantage. Datastore-based method exhibits lower reception lengths, primarily because they rely on the distribution of the database, which does not align with the model’s distribution. Draft model based method achieves better alignment with the original LLM due to dedicated training, but their smaller parameter size limits their ability to fully match the original model. In comparison, draft tokens from our method are consistently sourced from the same LLM, ensuring a perfectly matched distribution, which results in a higher acceptance rate.
\paragraph{Draft Tokens Construction Latency}
\begin{figure}[ht]
\centering
\includegraphics[width=0.9\linewidth]{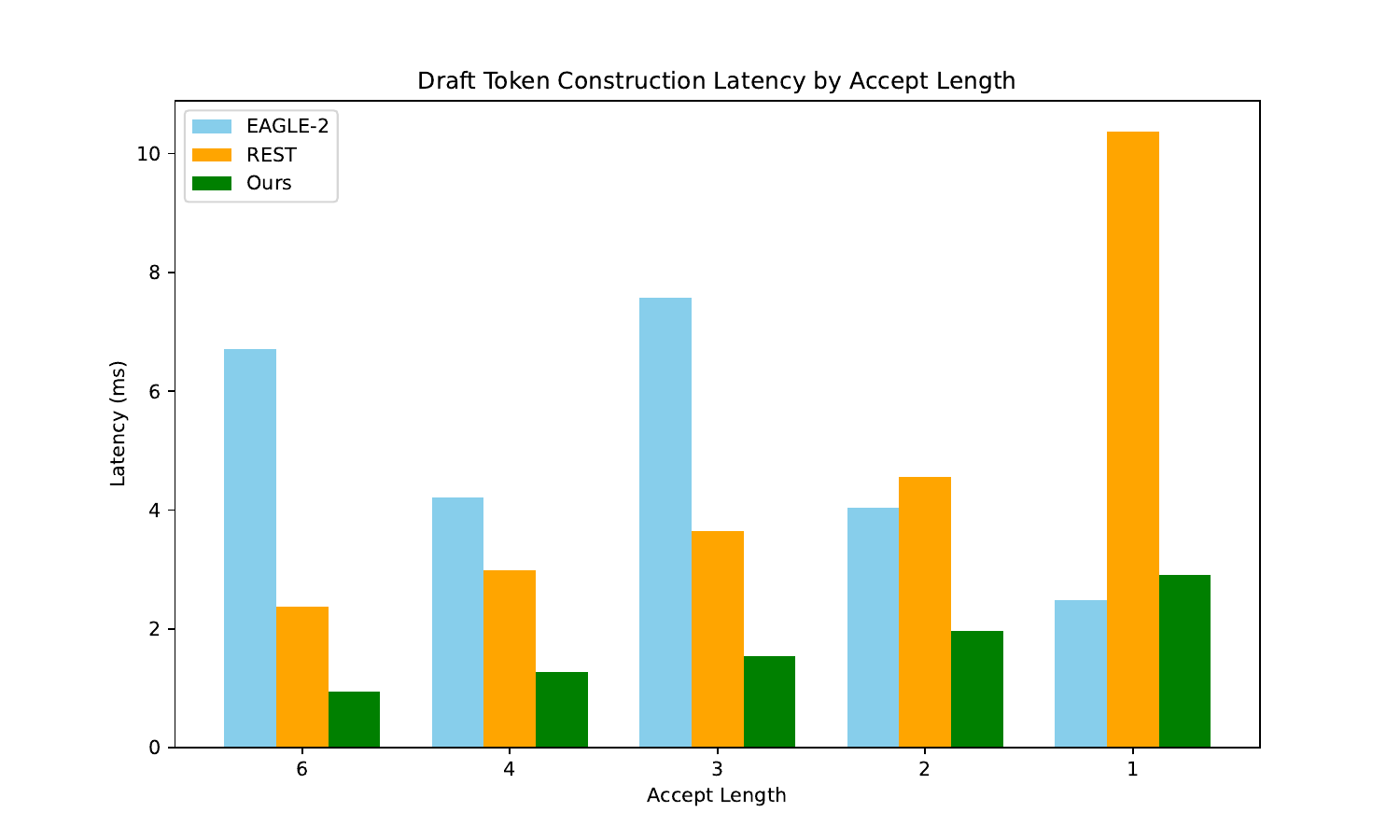}
\caption{Latency (ms) of draft token construction by Accept Length on Qwen2.5-7B-Instruct for GSM8K. Our method generally requires less time to construct draft tokens.}
\label{fig:latency}
\end{figure} 
Figure~\ref{fig:latency} reports the time consumption of different methods for constructing draft tokens. Among them, EAGLE-2 incurs the highest computational cost due to its reliance on GPU-based inference during construction. Our method is more efficient than REST, as our candidate set is significantly smaller than that of large-scale databases.

\subsection{Analysis}
\paragraph{Ablation Study}
\begin{figure}[ht]
\centering
\includegraphics[width=0.9\linewidth]{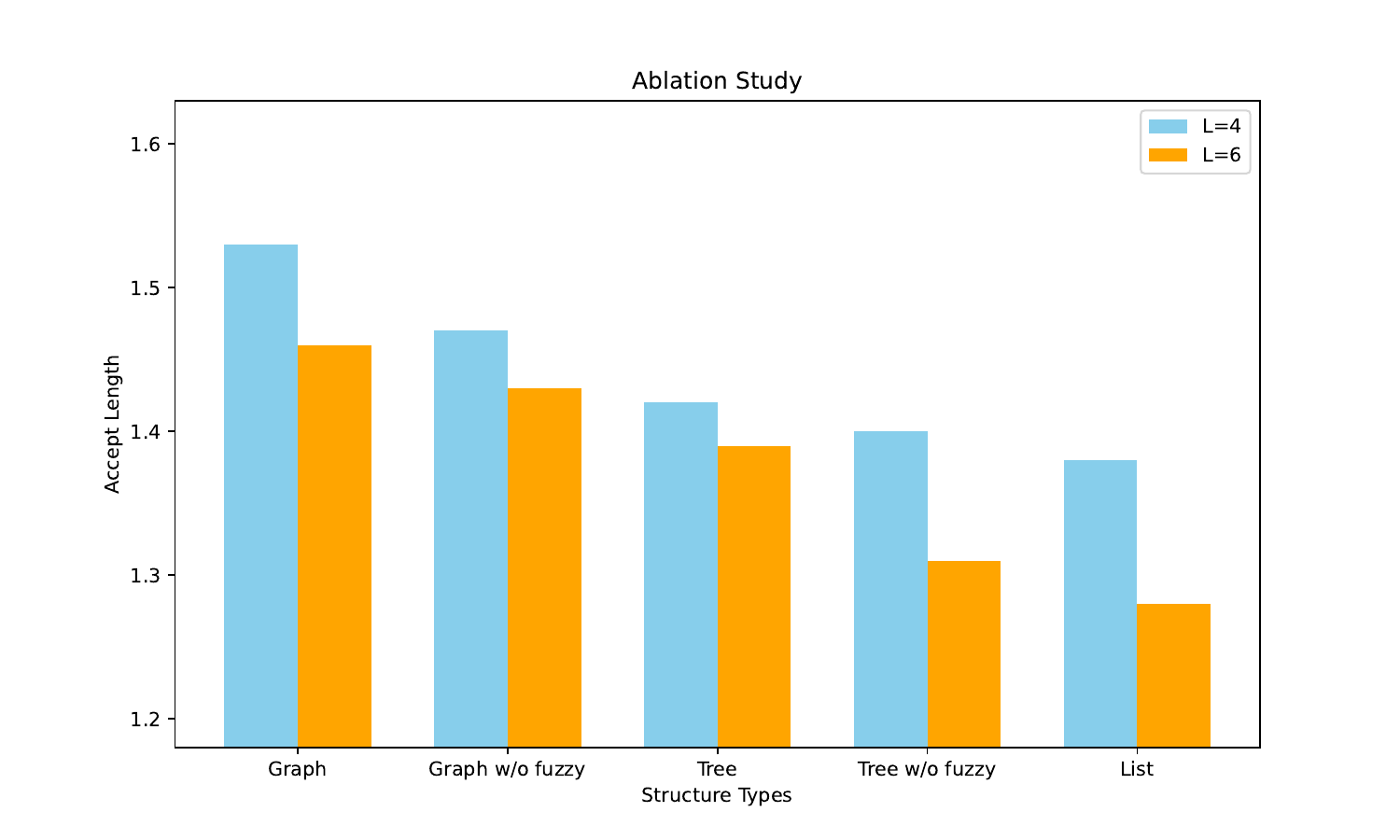}
\caption{Results of ablation study on Qwen2.5-7B-Instruct for GSM8K. DAG data structure with fuzzy matching achieved the best performance.}
\label{fig:analysis_num}
\end{figure} 
Figure 3 presents the results of the ablation study, demonstrating the effectiveness of different components of our method. It can be observed that, in modeling the data structure of the draft pool, DAG outperforms the tree, which in turn outperforms the list. This indicates that the DAG better captures consensus information among samples, leading to a higher acceptance rate. Additionally, we find that even when using a list, our method still achieves competitive performance, confirming that our approach generates high-quality draft tokens. Removing fuzzy matching leads to a performance drop, highlighting its role in enhancing the robustness of the draft pool.
\paragraph{Temperature Affection}
\begin{figure}[h]
    \centering
    \begin{minipage}{0.23\textwidth}
        \centering
        \includegraphics[width=\textwidth]{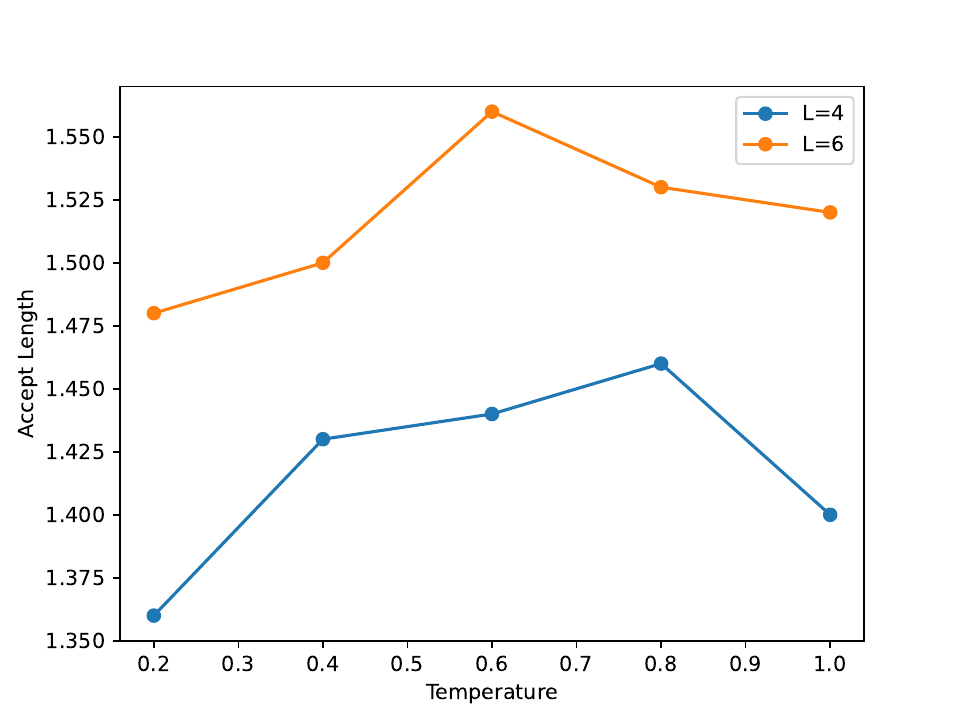}
    \end{minipage}
    \begin{minipage}{0.23\textwidth}
        \centering
        \includegraphics[width=\textwidth]{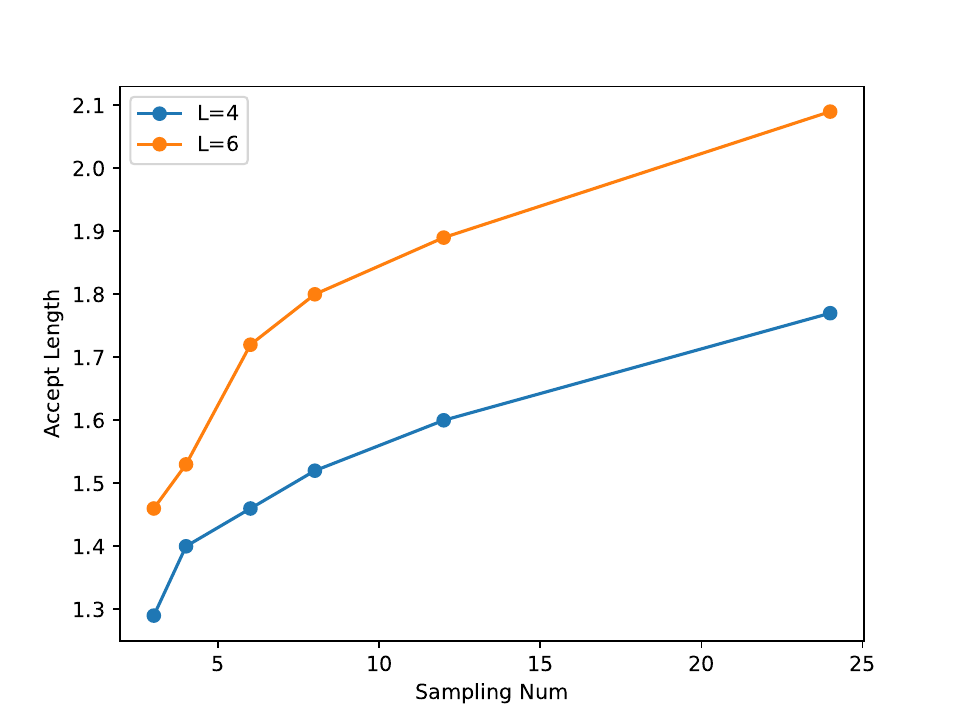}
    \end{minipage}
    
    \caption{The effect of sampling temperature (left) and size (right) on accept length on Qwen2.5-7B-Instruct for GSM8K. The optimal acceptance length occurs at intermediate temperature values and with a large parallel sampling size.}
    \label{fig:comparison}
\end{figure}

We investigate the impact of temperature on accept length under the same draft length. As shown in Figure~\ref{fig:comparison} (left), accept length exhibits a trend of first increasing and then decreasing as temperature rises. This is because, at lower temperatures, the diversity among samples is low, resulting in an insufficient number of candidates in the draft pool. Conversely, at higher temperatures, the diversity among samples increases, reducing the success rate of suffix matching. Therefore, the optimal temperature lies in the middle range.
\paragraph{Sampling Number Affection}
We further examine the effect of sampling size on accept length under the same draft length. As shown in Figure~\ref{fig:comparison} (right), accept length increases as the sampling size grows. This is because a larger parallel sampling size provides a richer set of candidates in the draft pool, leading to a higher acceptance rate for the selected draft tokens.

%% file: related_work.tex
\section{Related Work}
\paragraph{Multi-Sample Inference}
Multi-sample inference strategies have become pivotal for enhancing the performance and reliability of LLMs in complex reasoning tasks. Techniques like self-consistency \citep{SC} and Best-of-N sampling \citep{bestn} leverage parallel generation of multiple candidate outputs to mitigate individual errors and improve accuracy in math reasoning and many other tasks \citep{usc,wang-etal-2024-integrate}. One line of work aims to improve the performance of multiple sampling methods, for example, through weighted voting \citep{li-etal-2023-making,TDG}, enhancing the diversity of reasoning chains \citep{PTSC,li2025revisiting}, or step-wise guidance \citep{beam}. However, the computational cost of generating multiple samples remains a critical bottleneck. Recent work has explored optimizing multi-sample efficiency on controlling the sampling size \citep{ESC,aggarwal-etal-2023-lets,DSC,RASC}, yet the fundamental trade-off between accuracy gains and costs persists. Our method aims to accelerate multiple sampling, significantly reducing inference latency while ensuring no performance degradation.

\paragraph{Speculative Decoding}
Speculative decoding \citep{spe,lookahead} has emerged as a key paradigm for accelerating LLM inference by reducing auto-regressive steps without altering output quality.
However, these methods still struggle to fully align drafts with the original model’s distribution \citep{zhang-etal-2024-draft,medusa} and require significant cost for training \citep{eagle,li-etal-2024-eagle,distillspe,HASS} or storage \citep{he-etal-2024-rest}. Our work diverges by exploiting the inherent redundancy in multi-sample outputs, obviating external resources while achieving superior alignment through consensus-driven drafts.

%% file: conclusion.tex
\section{Conclusion}
We present a speculative decoding method that leverages consensus patterns from parallel reasoning paths to accelerate multi-sample inference, eliminating external dependencies. By dynamically synthesizing draft tokens through probabilistic aggregation and graph-based selection, our approach achieves higher acceptance rates and lower latency than baselines on mathematical reasoning benchmarks. This work bridges speculative decoding with sampling-based reasoning, establishing a resource-efficient paradigm for accelerating LLMs while preserving their output distribution. This approach could pave the way for the implementation of efficient multi-sample strategies in real-world applications.

%% file: limitations.tex
\section*{Limitations}
Our method has three key limitations:
\begin{itemize}
    \item Our method is not combined with speculative decoding methods optimized for batch processing, so its acceleration effect on time requires further exploration.
    \item The probabilistic aggregation mechanism incurs non-negligible overhead when processing a large number of parallel paths, potentially diminishing latency gains in extreme-scale scenarios.
    \item While validated on mathematical reasoning, its generalization to open-ended generation tasks remains unproven.
\end{itemize}

\section*{Ethics Statement}
All of the datasets used in this study were publicly available, and no annotators were employed for our data collection. We confirm that the datasets we used did not contain any harmful content and was consistent with their intended use (research). We have cited the datasets and relevant works used in this study.

%% file: appendix.tex
\newpage
\appendix

\section{Algorithm}
\label{app:agm}

\begin{algorithm}[ht]
\caption{Self-Consistency Draft Token Generation}
\begin{algorithmic}[1]
\State \textbf{Input:} Reasoning paths $\{y_1, y_2, \dots, y_N\}$ at $t$ step, suffix length $k$, maximum draft token length L, tolerance $\epsilon$
\State \textbf{Output:} Optimal draft token sequence $\mathcal{D}$
\State \textbf{Initialization:} 
\State \quad Construct the initial draft pool $\mathcal{C}_0 \gets \emptyset$
\For{each path $y_i$ in $\{y_1, y_2, \dots, y_N\}$}
    \State Find the $k$-token suffix $y_i^{t-k+1:t}$
    \For{each path $y_j$ where $j \neq i$}
        \State Search for fuzzy matching prefixes in $y_j$ and add subsequent tokens to candidate pool $\mathcal{C}_t$
    \EndFor
\EndFor
\State Construct Directed Acyclic Graph (DAG) $\mathcal{G} = (V, E)$:
\For{each token pair $u, v$ in the candidate pool}
    \State Compute edge weight $w(u, v) \gets \alpha \cdot p_\theta(v | u) + (1 - \alpha) \cdot \text{Freq}(u \to v)$
\EndFor

\State Greedily select tokens for optimal sequence:
\State \quad Initialize current sequence $\mathcal{D} \gets \emptyset$
\While{current draft token length $len(\mathcal{D})$ < $L$ and DAG is not empty}
    \State Select token $v_i$ with the highest weight $\sum_{u : (u, v) \in E} w(u,v)$ at each layer
    \State Add token $v_i$ to the draft token sequence $\mathcal{D}$
\EndWhile

\State \Return $\mathcal{D}$
\end{algorithmic}
\end{algorithm}

\section{Details about Experimental Settings}

\begin{figure}[ht]
\centering
\includegraphics[width=\linewidth]{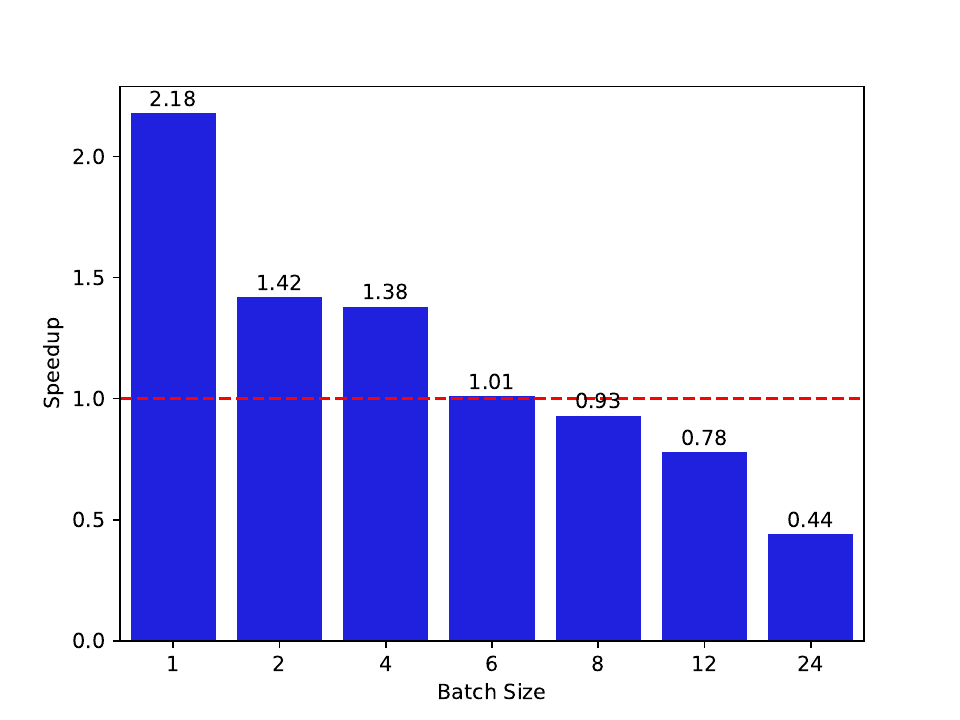}
\caption{Speedup ratio with different batch size.}
\label{fig:speedup}
\end{figure}

The raw evaluation setting for EAGLE-2 and REST both use a batch size of 1, and the inherently sequential execution of LLM leads to low hardware utilization on modern GPUs under this setting. Thus, the draft token length can be set quite large (e.g., 60). \citet{synergy} find that as the batch size increases, the optimal draft token length decreases. We reproduce BASS \citep{bass} multi-sample speculative decoding method on EAGLE-2 and find that when the draft token length is fixed to 4, the speedup relative to vanilla decoding gradually diminishes, and when the batch size is greater than or equals to 8, it even becomes slower than vanilla decoding. Detailed results of this experiment is shown in Fig~\ref{fig:speedup}. 
In this work, we focus on constructing draft tokens with higher acceptance rates, and accelerating parallel draft token verification and draft token generation are orthogonal tasks. Therefore, how to accelerate parallel draft token verification is not within the scope of our study, and different draft token construction methods can be validated using the same approach. As a result, we adopt accept length as the evaluation metric for speculative decoding methods, rather than speedup. Additionally, we keep the computational load of a single multi-sample speculative decoding step constant, i.e., draft token length * batch size = 24. Furthermore, since draft token length = draft sequence length * draft sequence number for tree decoding in EAGLE-2 and REST, when the draft token length is fixed, there are different schemes to generate draft candidates (e.g., when the draft token length is 4, we can either generate 2 draft sequences of length 2 or 1 draft sequence of length 4). In this work, both EAGLE-2 and REST report the results of the best draft candidates generation scheme under the specified draft token length.

\label{app:metric}